\documentclass[letterpaper]{article} 

\usepackage{aaai24}  

\usepackage{times}  
\usepackage{helvet}  
\usepackage{courier}  
\usepackage[hyphens]{url}  
\usepackage{graphicx} 
\usepackage{natbib}  
\usepackage{caption} 
\usepackage{algorithm}
\usepackage{algorithmic}

\usepackage{newfloat}
\usepackage{listings}
\DeclareCaptionStyle{ruled}{labelfont=normalfont,labelsep=colon,strut=off} 
\floatstyle{ruled}
\newfloat{listing}{tb}{lst}{}
\floatname{listing}{Listing}
\usepackage{amsmath,amssymb,bm}
\newcommand{\name}{WeditGAN}
\newcommand{\dw}{\Delta w}
\newcommand{\wsrc}{w_\mathrm{src}}
\newcommand{\wtgt}{w_\mathrm{tgt}}
\newcommand{\W}{\bm{W}}
\newcommand{\Wp}{\bm{W}^+}
\newcommand{\Wpsrc}{\bm{W}_{\mathrm{src}}^+}
\newcommand{\Wptgt}{\bm{W}_{\mathrm{tgt}}^+}
\newcommand{\Isrc}{I_{\mathrm{src}}}
\newcommand{\Itgt}{I_{\mathrm{tgt}}}
\newcommand{\supp}{Appendix}
\newcommand{\Lperp}{L_{\mathrm{perp}}}
\newcommand{\E}{\mathbb{E}}

\usepackage{xspace}
\makeatletter
\DeclareRobustCommand\onedot{\futurelet\@let@token\@onedot}
\def\@onedot{\ifx\@let@token.\else.\null\fi\xspace}

\def\eg{\emph{e.g}\onedot} 
\def\ie{\emph{i.e}\onedot} 
 
\def\etc{\emph{etc}\onedot} 
\def\wrt{w.r.t\onedot}

\makeatother

\usepackage[capitalize]{cleveref}
\Crefname{section}{Section}{Sections}
\Crefname{table}{Table}{Tables}
\Crefname{figure}{Figure}{Figures}

\title{WeditGAN: Few-Shot Image Generation via Latent Space Relocation}
\author{
    Yuxuan Duan\textsuperscript{\rm 1}, Li Niu$^{*}$\textsuperscript{\rm 1}, Yan Hong\textsuperscript{\rm 2}, Liqing Zhang\thanks{Corresponding authors.}\textsuperscript{\rm 1}\\
}
\affiliations{
    \textsuperscript{\rm 1}MoE Key Lab of Artificial Intelligence, Shanghai Jiao Tong University \\
    \textsuperscript{\rm 2}Tiansuan Lab, Ant Group\\
    sjtudyx2016@sjtu.edu.cn, ustcnewly@sjtu.edu.cn, yanhong.sjtu@gmail.com, zhang-lq@cs.sjtu.edu.cn
}

\begin{document}

\maketitle

\begin{abstract}
In few-shot image generation, directly training GAN models on just a handful of images faces the risk of overfitting. A popular solution is to transfer the models pretrained on large source domains to small target ones. In this work, we introduce \name{}, which realizes model transfer by editing the intermediate latent codes $w$ in StyleGANs with learned constant offsets ($\Delta w$), discovering and constructing target latent spaces via simply relocating the distribution of source latent spaces. The established one-to-one mapping between latent spaces can naturally prevents mode collapse and overfitting. Besides, we also propose variants of \name{} to further enhance the relocation process by regularizing the direction or finetuning the intensity of $\dw$. Experiments on a collection of widely used source/target datasets manifest the capability of \name{} in generating realistic and diverse images, which is simple yet highly effective in the research area of few-shot image generation. Codes are available at \url{https://github.com/Ldhlwh/WeditGAN}.
\end{abstract}

\section{Introduction}
\label{sec:intro}


While achieving convincing performance in many tasks since proposed by \citet{gan}, Generative Adversarial Networks (GANs) are renowned for the enormous data they require to possess satisfying fidelity and variety of the generated samples. Mainstream datasets for image generation methods typically cover 20k (CelebA) \cite{celeba}, 70k (FFHQ) \cite{stylegan1} or 126k (LSUN church) \cite{lsun} items. 
Some researches focus on efficient data usage by enhancing the training processes of GANs with differentiable augmentation \cite{diffaug, ada}. These methods lower the data threshold to hundreds or thousands, yet leaving training GANs on even fewer images (\eg ten or five) unsolved. Another paradigm of solutions to \textbf{few-shot image generation (FSIG)} is model transfer, where GANs pretrained on source domains with large datasets are adapted to target domains with only a handful of images. Finetuning-based methods reduce the number of trainable parameters trying to alleviate the overfitting issue \cite{tgan, bsa, svd, freezed, minegan}, yet the effects are usually limited. Other methods propose regularization terms imposing penalties on feature/parameter changes during the transfer processes \cite{ewc, cdc, rssa, dcl, dwsc}. By minimally updating the model, these methods seek to keep some characteristics of the source images, hence inherit the diversity of the source domain. Nevertheless, regularization-based methods often face the dilemma of balancing the characteristics of the source/target domains, since sacrificing the characteristics of the target domain impairs the image fidelity.


\begin{figure}[t]
  \includegraphics[width=\linewidth]{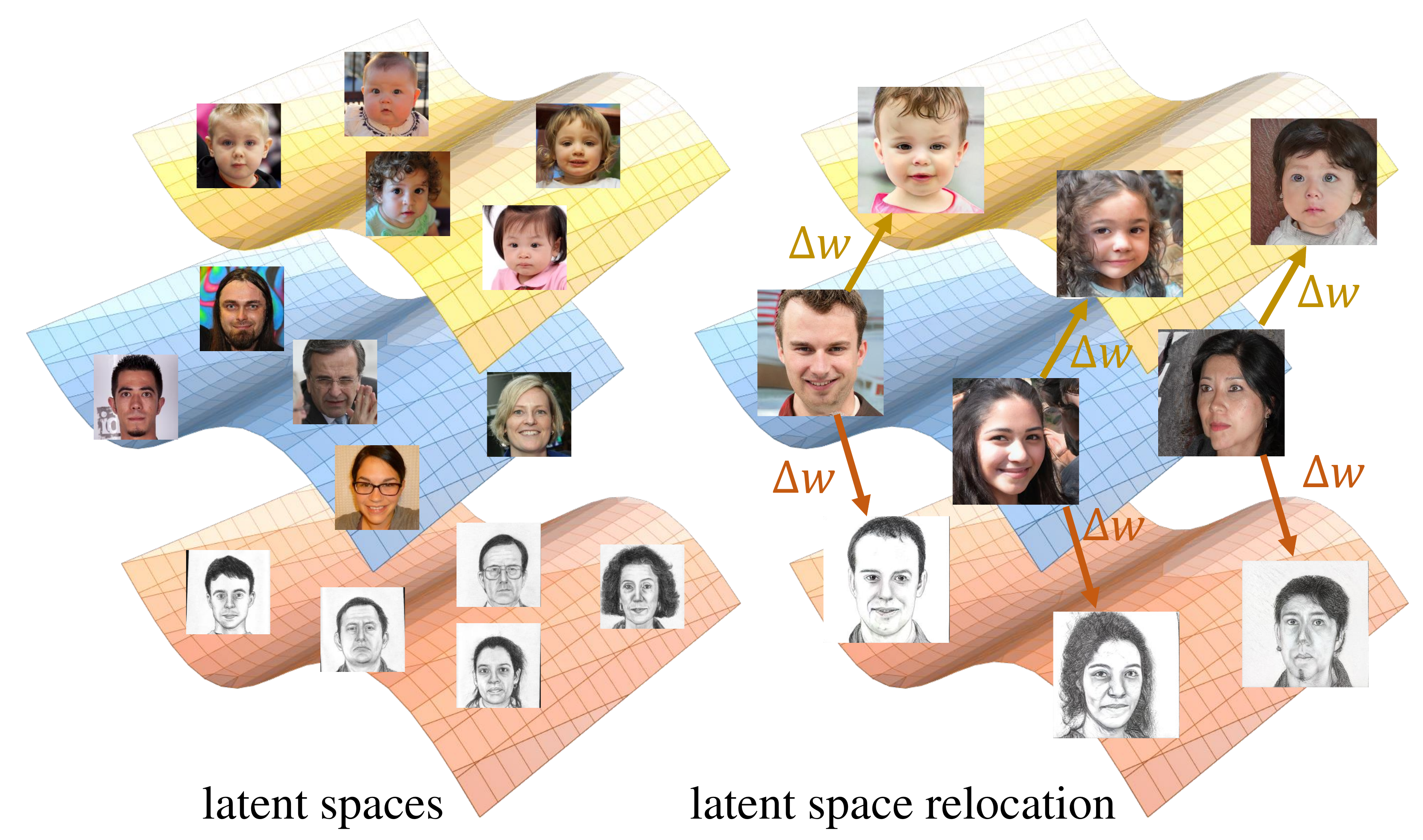}
  \caption{The core idea of latent space relocation with constant latent codes $\dw$, based on the fact that the latent spaces of related domains in the same generative model share similar shapes of manifolds.}
  \label{fig:teaser_2}
\end{figure}

StyleGANs \cite{stylegan1, stylegan2} are among the most popular GANs nowadays. They not only manifest strong generative abilities, but also construct an intermediate latent space $\W$. This latent space is highly disentangled and well aligned to dataset attributes, which enables a variety of latent space manipulation methods for image editing \cite{editinginstyle, ganalyze, ganspace}. Most of these methods focus on in-domain manipulation, such as editing the expression, appearance and/or facial pose of a face into another face. However, since some previous works discover that out-of-distribution latent codes may still render reasonable images \cite{image2stylegan, image2styleganpp, e4e}, we suppose that cross-domain model transfer in FSIG tasks can also be achieved by proper latent space manipulation.

According to \citet{dataauggan}, the latent spaces of related domains (such as face photos and artistic portraits) in the same model have similar shapes of manifolds, such as the disentangled linear latent space $\W$ of StyleGANs \cite{interpreting}. Previous work StyleCLIP \cite{styleclip} has validated that for in-domain manipulation, the editing directions are nearly identical when editing the same attribute on different facial images. 
In preliminary experiments of our work, we find that such identity can be extended to cross-domain scenarios, where the average pairwise cosine similarity of the adaptive editing offsets on different images is over $98\%$ (refer to \supp{} for details). 
Inspired by this idea, we propose \name{} ($w$-edit GAN), which relocates the latent space $\W$ of the target domain from that of the source domain by learning a constant latent offset $\dw$ when transferring the pretrained model to the few-shot target domain, as shown in \cref{fig:teaser_2}. 
Although such constant $\dw$ implies exactly identical shapes of manifolds between latent spaces, which seems restrictive and may not be absolutely true, we believe it is an adequately good approximation in FSIG, where learning extra modules giving adaptive $\dw$ based on each source latent code are prone to overfitting (see \cref{sec:ablation}).
Instead, using constant $\dw$ naturally establishes a one-to-one mapping between source/target latent codes, perfectly keeps the shape of the source latent space, and eventually prevents overfitting by inheriting the diversity from the source dataset.

Later in \cref{sec:experiment}, we compare \name{} with several state-of-the-art works on a variety of source/target dataset pairs. Although \name{} is a finetuning-based method, experiment results show that it is actually good at maintaining the diversity from the pretrained model as the recent regularization-based ones. Besides, without strict penalties on parameter/feature changes during model transfer, \name{} as well outperforms recent methods in capturing characteristics of the target domain, producing images with high fidelity without loss of diversity. Last but not least, We also propose variants of \name{}, including a perpendicular regularization on $\dw$ for a more precise relocation, finetuning the editing intensity for better quality, and equipping \name{} with contrastive loss to show the orthogonality of our work and the regularization-based methods. 

Our contributions can be summarized as follows: 
(1) We design the simple yet highly effective \name{} which transfer the model from the source to the target domain by relocating the latent space with learned constant $\dw$; 
(2) We explore several variants of \name{}, which further enhance the performance and indicate possible research directions for future works; 
(3) Extensive experiments on commonly used pairs of source/target datasets for FSIG verify the state-of-the-art performance of \name{}.
\section{Related Work}
\label{sec:related}

\paragraph{Latent Space Manipulation}

Manipulating the latent space of StyleGAN was mostly done on the extended latent space $\Wp$ \cite{image2stylegan} or style space $\bm{S}$ \cite{styleintervention, stylespace}. GAN inversion methods \cite{image2stylegan, image2styleganpp, e4e, psp} provided latent codes of arbitrary in-domain image for editing. Among the manipulation methods, some fused two latent codes to combine certain attributes from the two images for semantic editing \cite{editinginstyle}, style transfer \cite{dualstylegan} or face reenactment \cite{reenact, stylerig}. Other methods edited a single latent code along certain directions either supervised \cite{interpreting, ganalyze, styleclip} or unsupervised \cite{ganspace}. Nevertheless, most of the previous works focused on in-domain manipulation, leaving cross-domain manipulation rarely researched. Besides, some works extracted latent codes or offsets by training extra encoders \cite{dualstylegan, stylerig}, which limited their applicability in few-shot scenarios since these encoders would render collapsed latent spaces given limited data. On the contrary, our \name{} proposes constant $\dw$ to bridge the source/target latent spaces without collapse, performing cross-domain manipulation.

\paragraph{Few-shot Image Generation}

In FSIG, the paradigm of model transfer methods can be divided into two types.

The finetuning-based methods reduce the number of trainable parameters by finetuning a part of the model \cite{freezed, leveraging}, training additional parameters while fixing the main model \cite{bsa, minegan} or decomposing the parameters \cite{svd}. Taking image quality and diversity into regard, earlier methods were generally not the most competitive. Nevertheless, two recent works \cite{adam, rick} probed the important parameters using Fisher Information \cite{fi} to use different finetuning policies accordingly, reaching comparable results with the regularization-based methods in recent years.

The regularization-based methods usually finetune all the model parameters but introduce penalties on parameter/feature changes and encouraging feature distribution alignment. EWC \cite{ewc} decided the strength of penalties via Fisher Information \cite{fi}. CDC \cite{cdc} maintained feature similarity among source/target samples with consistency loss. RSSA \cite{rssa} introduced spatial consistency losses to keep structural identity. DCL \cite{dcl} proposed contrastive losses between source/target features of both generator and discriminator. DWSC \cite{dwsc} designed perceptual/contextual loss respectively for easy/hard-to-generate patches. These methods kept the characteristics of the source domain by imposing strong regularization thus inherited the diversity from the source dataset. Yet they were prone to lose characteristics of the target domain, especially when the characteristics of the two domains are in conflict (see \cref{sec:fsig}).

Another paradigm of FSIG adopt the idea of \emph{seen categories to unseen categories}. Methods of this paradigm trained an image-to-image model on some categories of a multi-class dataset, and then directly test the model on the other categories \cite{f2gan, matchinggan, deltagan, lofgan, age}. Besides, there are also some methods for specialized FSIG scenarios, such as font generation \cite{font1, font2} and defect generation \cite{dfmgan}. Since the definition to the tasks and/or the experimental settings of these works significantly differ from ours thus less relevant, we will not detail them in this work.

\paragraph{Few-shot Domain Adaptation}

As a task related to FSIG, the essential distinction between the two tasks is that few-shot domain adaptation (FSDA) requires that the two images before/after adaptation ideally have the same content (\eg depicting the same person), where in FSIG tasks the fidelity to the target domain and the diversity are the only metrics to evaluate the generated images. Recent works of FSDA usually rolled out in one-shot scenarios \cite{jojogan, generalizedosda, mindthegap, stylegannada, essence, towardsosda, justoneclip}. Their methods captured the characteristics of a single target image using encoders, instead of a target domain whose distribution is learned by discriminators in image generation tasks. Most of these methods cannot be directly adapted to few-shot scenarios either without non-trivial redesigns.

\begin{figure*}[t]
  \centering
  \includegraphics[width=\linewidth]{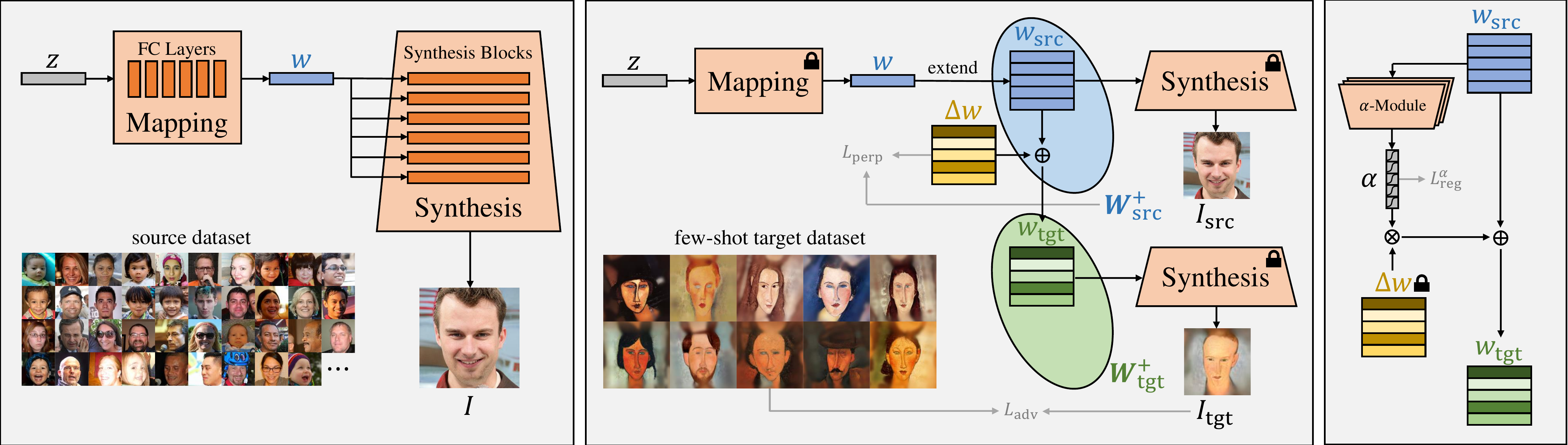}
   \caption{The procedure of \name{}. Left: A StyleGAN is first trained on a large source dataset. Middle: During the transfer process, the mapping and the synthesis network are both fixed. With the target latent code $\wtgt$ constructed by summing up $\wsrc$ and the only trainable parameters $\dw$, the synthesis network can generate images of the target domain. Right: After $\dw$ is learned and fixed, \name{} trains a set of AlphaModules to finetune the editing intensity customized for each $\wsrc$ (optional).}
   \label{fig:weditgan}
\end{figure*}

\section{Method}
\label{sec:method}

\name{} follows the common noise-to-image pipeline of image generation tasks. First, we pretrain a StyleGAN model on a large source domain with abundant data. Then, we transfer the model via latent space relocation to a relevant small target domain containing no more than ten images.

\subsection{StyleGAN Preliminary}

\name{} adopts the most widely used StyleGAN2 \cite{stylegan2} as its base model. The generator of StyleGAN consists of a mapping network $M$ and a synthesis network $S$. The mapping network maps a random noise code $z$ to a latent code $w$ in the intermediate latent space $\W$. Then the synthesis network, whose synthesis blocks are made up of convolutional layers, takes $w$ and produces an image $I$. The whole process can be formulated as
\begin{equation}
    z \sim \mathcal{N}(0, \mathbf{I}), \quad w = M(z), \quad I = S(w),
\end{equation}
where $w$ is transformed by the learned affine layers and modulates the weights of the filters of each convolutional layers in the synthesis network. We follow the normal procedure of pretraining StyleGANs on source datasets. 

\subsection{\name{}}

During model transfer from source to target domains, \name{} mainly works on the latent space. It seeks an appropriate constant $\dw$ which can relocate the latent space of the target domain from the source latent space by bridging the gap between these two distributions.

\paragraph{Extended Latent Space}

StyleGAN with default settings uses the same $w \in \W \subseteq \mathbb{R}^{512}$ to modulate every convolutional layer. 
However, to increase flexibility, we conduct latent space relocation on the extended latent space $\Wp \subseteq \mathbb{R}^{n \times 512}$ \cite{image2stylegan}, which uses separate latent codes for each of the $n$ convolutional layers. Practically, \name{} makes $n$ copies of the original $w$ into $\wsrc \in \Wpsrc$, and then edits $\wsrc$ in a layer-wise manner, obtaining $\wtgt \in \Wptgt$ with possibly different latent codes modulating each layer respectively. Ablation study in \cref{sec:ablation} demonstrates the necessity of using $\Wp$.

\paragraph{Latent Space Relocation}

The fundamental idea of \name{} is to find the target latent space $\Wptgt$ based on the source space $\Wpsrc$ with a learned constant latent offset $\dw \in \mathbb{R}^{n \times 512}$, so that the StyleGAN pretrained on the source domain can generate target domain images by merely taking target latent codes $\wtgt$ instead of $\wsrc$. As shown in \cref{fig:weditgan}, we introduce new parameters $\dw$ to the generator. During the model transfer processes, we only update $\dw$ by training on a few target images, yet fix both the mapping and the synthesis network. Since the parameters of these two networks remain unchanged, \name{} still keeps the ability of generating source domain images which previous works fail. \name{} is capable of producing paired source/target images using a single model by simply switching between $\wsrc$ and $\wtgt$. The generation process can be formulated as
\begin{equation}
    \begin{aligned}
        & z \sim \mathcal{N}(0, \mathbf{I}), \quad w = M(z), \quad \wsrc = \mathrm{extend}(w), \\
        & \wtgt = \wsrc + \dw, \quad \Isrc = S(\wsrc), \quad \Itgt = S(\wtgt).
    \end{aligned}
    \label{eq:lsr}
\end{equation}
In total, only $512 \times n$ parameters need training. Since $n$ is typically $20$ for StyleGAN2 on $256^2$, $\dw$ is only $0.04\%$ of the generator parameters. Compared with the previous methods which finetune the whole model or a large portion of its parameters, \name{} is less sensitive to data insufficiency. See \supp{} for details.


\paragraph{Objective}

Besides the generator $G$, \name{} also finetunes the pretrained discriminator $D$ without special strategy, following recent works \cite{ewc, cdc, rssa}. Ablation study in \cref{sec:ablation} also manifests that the overfitting issue of the discriminator will be alleviated when the diversity of the generator is assured.

Similar to common GAN models, \name{} alternatively updates $G$ and $D$ via the original losses of StyleGAN:
\begin{equation}
    L(G, D) = L_{\mathrm{adv}}(G, D) + L_{\mathrm{pl}}(G) + L_{\mathrm{R1}}(G, D),
    \label{eq:obj}
\end{equation}
where $L_{\mathrm{adv}}$, $L_{\mathrm{pl}}$, $L_{\mathrm{R1}}$ respectively represent adversarial loss, path length regularization encouraging smooth and disentangled latent spaces, and R1 regularization stabilizing the training process by adding gradient penalty. Refer to \citet{stylegan1} for details.



\subsection{\name{} Variants}

\paragraph{Perpendicular Regularization}

In \name{}, the latent offset $\dw$ is learned in a fully data-driven manner. The discriminator provides supervision to $\dw$ so that the relocated $\Wptgt$ aligns with the target dataset. However, there might be a gap between the dataset distribution $P_{\mathrm{data}}$ and the domain distribution $P_\mathrm{domain}$ due to the potentially biased samples in few-shot datasets \cite{survey}. As a result, \name{} may relocate $\Wptgt$ where the distribution of the generated images $P_\mathrm{gen}$ is closed to $P_{\mathrm{data}}$ instead of the expected $P_\mathrm{domain}$, by unnecessarily editing in-domain attributes in the source latent space. In order to restrain such unnecessary edits and enhance diversity, we would like to make $\dw$ perpendicular to the manifold of the source latent space. For instance, the learned $\dw$ in \name{} transferring from facial photos to sketches should be perpendicular to the editing offset from male to female in the source latent space, as altering the gender is not necessary between the two domains.

It is non-trivial to obtain the closed-form representation of $\Wpsrc$, and we can only sample $\wsrc$ by random input $z$ and the mapping network $M$. Since $M$ is a continuous function, the perpendicular regularization between $\dw$ and $\Wpsrc$ can be formulated as below with a small $\sigma \to 0$:
\begin{equation}
    \small
    \Lperp = \E_{z \sim \mathcal{N}(0, \mathbf{I}), \varepsilon \sim \mathcal{N}(0, \sigma^2 \mathbf{I})}[\langle \dw, M(z + \varepsilon) - M(z) \rangle^2].
    \label{eq:perp}
\end{equation}
For brevity, \cref{eq:perp} can be approximated by relaxing the neighborhood requirement and using pairs of random $\wsrc$ since $\Wpsrc$ is a highly disentangled and linear manifold:
\begin{equation}
    \Lperp = \E_{\wsrc^1, \wsrc^2 \sim P(\wsrc)} [\langle \dw, \wsrc^1 - \wsrc^2 \rangle^2].
    \label{eq:perp1}
\end{equation}
During model transfer, for a batch $\{\wsrc^i\}_{i=1}^m$, the perpendicular loss in \cref{eq:perp1} can be computed on-the-fly by
\begin{equation}
    \Lperp = \sum_{1 \le i,j \le m, i \neq j} \langle \dw, \wsrc^i - \wsrc^j \rangle^2.
    \label{eq:perp2}
\end{equation}
We choose the inner product $\langle \cdot, \cdot \rangle$ instead of $\cos(\cdot, \cdot)$ because the former can also decrease the magnitude of $\dw$ to avoid excessive edits. The perpendicular loss in \cref{eq:perp2} are appended to the original objective function \cref{eq:obj} with a hyperparameter weight $\lambda_{\mathrm{perp}}$.

\paragraph{Editing Intensity Finetuning}

\name{} relocates the whole latent space with a constant $\dw$, which is learned to render good images for most $\wsrc \in \Wpsrc$. Nevertheless, some $\wsrc$ at the margin of the distribution of $\Wpsrc$ may still be relocated to suboptimal target images. For example, when transferring the facial photo domain to babies, source latent codes $\wsrc$ corresponding to the elders/youths need more/less edits along the constant $\dw$ learned for the whole source domain. Hence, a customized editing intensity $\alpha$ for each $\wsrc$ may further improve the overall image quality.

After the constant $\dw$ is learned and fixed, an additional lightweighted \emph{AlphaModule} with two FC layers is attached to each synthesis block. These modules are learned to finetune the editing intensity based on each $\wsrc$. The latent space relocation in \cref{eq:lsr} is now formulated as 
\begin{equation}
   \wtgt = \wsrc + [1 + \mathrm{AlphaModule}(\wsrc)] \cdot \dw,
\end{equation}
where the output of AlphaModules are activated by $\tanh$ to provide intensity residuals $\alpha$ in $(-1, 1)$ (weakened to strengthened), resulting in customized editing intensity in $(0, 2)$. We add L2 regularization (weighted by hyperparameter $\lambda_{\mathrm{reg}}^{\alpha}$) to discourage significantly deviated intensities.

\paragraph{Orthogonality to Regularization-based Methods}

As a finetuning-based method, \name{} is orthogonal to the regularization-based methods adding penalties on feature changes. To verify that \name{} can be combined with such methods, we introduce another variant equipped with multilayer feature contrastive losses on both $G$ and $D$, which pull closer the features generated by the same $z$ and push apart those generated by different $z$. Akin to DCL \cite{dcl}, with a batch of latent codes $\{\wsrc^i, \wtgt^i\}_{i=1}^m$ and the corresponding generated images $\{\Isrc^i, \Itgt^i\}_{i=1}^m$, the contrastive losses can be formulated as:
\begin{equation}
    \begin{aligned}
        & L_{\mathrm{CL}}^G = \sum_{l}\sum_{i=1}^m -\log \frac{\phi[S^l(\wtgt^i), S^l(\wsrc^i)]}{\sum_{j=1}^m \phi[S^l(\wtgt^i), S^l(\wsrc^j)]}, \\
        & L_{\mathrm{CL}}^D = \sum_{l}\sum_{i=1}^m -\log \frac{\phi[D^l(\Itgt^i), D^l(\Isrc^i)]}{\sum_{j=1}^m \phi[D^l(\Itgt^i), D^l(\Isrc^j)]},
    \end{aligned}
\end{equation}
where $S^l(\cdot)$, $D^l(\cdot)$ is the feature of the $l$-th synthesis/discriminator block, and $\phi(\cdot, \cdot) = \exp(\mathrm{CosineSimilarity}(\cdot, \cdot))$. These contrastive losses are appended to the original objective function \cref{eq:obj} with a hyperparameter weight $\lambda_{\mathrm{CL}}$.

\section{Experiment}
\label{sec:experiment}

To validate the ability of \name{}, we conduct experiments on an extensive set of eight commonly used source/target dataset pairs. We detail the experimental settings in \cref{sec:setting}, illustrate and analyze results in \cref{sec:fsig}, and investigate \name{} through ablation study in \cref{sec:ablation}.

\subsection{Experimental Setting}
\label{sec:setting}

\paragraph{Dataset}

Following previous works, we mainly focus on model transfer from face photos to artistic portraits, including \textbf{FFHQ} $\to$ \textbf{Sketches} \cite{sketches}, \textbf{Babies}, \textbf{Sunglasses}, paintings by \textbf{Amedeo} Modigliani, \textbf{Raphael}, and \textbf{Otto} Dix \cite{faceofart}. We also test \name{} on \textbf{LSUN Church} $\to$ \textbf{Haunted} houses, and \textbf{LSUN Car} $\to$ \textbf{Wrecked} cars. All the target datasets contain only ten training images, with resolution of $256^2$.

\paragraph{Baseline}

We include the regularization-based methods CDC \cite{cdc}, DCL \cite{dcl}, RSSA \cite{rssa}, and DWSC \cite{dwsc}. We also include the latest finetuning-based methods AdAM \cite{adam} and RICK \cite{rick}.
Our experiments ensure a fair comparison by transferring from the same pretrained StyleGAN models. See \supp{}.

\paragraph{Metric}

We evaluate both quality and diversity of the generated images. For the three domains (Sketches, Babies, Sunglasses) sampled from larger full datasets, we calculate FID \cite{fid} between 5,000 generated samples and the full datasets. For other domains only containing ten images, we compute KID \cite{kid} instead, which is more precise than FID in few-shot cases \cite{ada}. 
Nevertheless, since ten images are probably insufficient to perfectly represent the target domains, such KID scores are shown for reference only.

We also report the intra-cluster version of LPIPS \cite{lpips} of 1,000 samples as a standalone diversity metric. 
For details, refer to \citet{cdc} where it originates. 



\subsection{Few-shot Image Generation}
\label{sec:fsig}

We conduct the experiments with \textbf{\name{}} and its three variants. \textbf{\name{} perp} impose perpendicular regularization, with $\lambda_\mathrm{perp} = 10^{-4}$. \textbf{\name{} alpha} finetunes the editing intensity after learning $\dw$, with $\lambda_{\mathrm{reg}}^{\alpha} = 1/0.1/0.01$ for different cases. \textbf{\name{} CL} appends contrastive losses on feature changes for multiple layers in both the generator and the discriminator, with $\lambda_{\mathrm{CL}} = 0.5$. Besides 10-shot, we also provide results of FSIG experiments in extreme 5-shot/1-shot settings in \supp{}.

\paragraph{Quantitative Result}

The quantitative results are listed in \cref{tab:quan1,tab:quan2}. Although our baselines consist of the most competitive recent methods, our \name{} and its variants still achieve state-of-the-art performance on both metrics. A possible reason is that finetuning the pretrained generator on few-shot datasets may inevitably harm the image quality and cause overfitting. Even though the previous works design specialized strategies trying to counteract such effect, they are still outperformed by \name{} which keeps the pretrained generator completely intact.

\begin{table}[t]
    \centering
    \small
    \tabcolsep=1.3mm{
    \begin{tabular}{lrr|rr|rr}
    \hline
     & \multicolumn{2}{c|}{Sketches} & \multicolumn{2}{c|}{Babies} & \multicolumn{2}{c}{Sunglasses} \\
    Method & FID$\downarrow$ & LPIPS$\uparrow$ & FID$\downarrow$ & LPIPS$\uparrow$ & FID$\downarrow$ & LPIPS$\uparrow$ \\
    \hline
    CDC & 70.65 & 0.4412 & 43.99 & 0.5859 & 34.77 & 0.5873 \\
    DCL & 57.72 & \textbf{0.4477} & 46.57 & 0.5833 & 31.37 & 0.5844 \\
    RSSA & 66.97 & \textbf{0.4448} & 55.50 & 0.5786 & 27.40 & 0.5748 \\
    DWSC & 61.03 & 0.4095 & 39.00 & 0.5604 & 31.20 & 0.5799 \\
    AdAM & 38.11 & 0.4446 & 42.44 & 0.5837 & 26.98 & 0.5957 \\
    RICK & 40.52 & 0.4310 & 39.41 & 0.5709 & 25.09 & 0.5992 \\
    \hline 
    \textbf{\name{}} & \textbf{35.41} & 0.4339 & \textbf{38.97} & 0.6174 & 21.72 & \textbf{0.6254} \\
    \textbf{+ perp} & 37.12 & \textbf{0.4504} & \textbf{37.78} & \textbf{0.6386} & \textbf{19.54} & \textbf{0.6424} \\
    \textbf{+ alpha}  & \textbf{34.13} & 0.4250 & \textbf{36.19} & \textbf{0.6296} & \textbf{17.06} & 0.6223 \\
    \textbf{+ CL} & \textbf{36.13} & 0.4176 & 40.22 & \textbf{0.6388} & \textbf{19.59} & \textbf{0.6472} \\
    \hline
  \end{tabular}}
  \caption{The results of FSIG on FFHQ $\to$ Sketches, Babies, and Sunglasses. We report FID@5k with the full datasets, and Intra-cluster LPIPS@1k. Top-3 results are in bold.}
    \label{tab:quan1}
\end{table}

\begin{table*}[t]
  \centering
    \small
  \begin{tabular}{lrr|rr|rr|rr|rr}
    \hline
    & \multicolumn{2}{c|}{Amedeo} & \multicolumn{2}{c|}{Raphael} & \multicolumn{2}{c|}{Otto} & \multicolumn{2}{c|}{Haunted} & \multicolumn{2}{c}{Wrecked}\\
    Method &  KID$\downarrow$ & LPIPS$\uparrow$ & KID$\downarrow$ & LPIPS$\uparrow$ & KID$\downarrow$ & LPIPS$\uparrow$ & KID$\downarrow$ & LPIPS$\uparrow$ & KID$\downarrow$ & LPIPS$\uparrow$ \\
    \hline
    CDC & 23.33 & 0.5860 & 8.58 & 0.5711 & 16.54 & 0.6579 & 23.96 & 0.6075 & 25.89 & 0.4571 \\
    DCL & 14.64 & 0.5756 & 3.66 & 0.5500 & 13.66 & 0.6480 & 27.69 & 0.6137 & 31.54 & 0.3928 \\
    RSSA & 20.24 & \textbf{0.6150} & 3.46 & 0.5545 & 18.74 & 0.6308 & 27.88 & 0.6011 & 27.13 & 0.3739 \\
    DWSC & 19.79 & 0.5515 & 22.63 & 0.5162 & 32.68 & 0.5924 & 24.45 & 0.5666 & 21.08 & \textbf{0.4883} \\
    AdAM & \textbf{12.73} & 0.5492 & 4.70 & 0.5744 & \textbf{9.89} & 0.6440 & 24.54 & 0.6228 & \textbf{16.34} & 0.3968 \\
    RICK & 18.12 & 0.5765 & 4.49 & 0.5569 & 11.24 & 0.6326 & 21.02 & 0.6201 & 35.73 & 0.3961 \\
    \hline 
    \textbf{\name{}} & \textbf{12.07} & 0.5874 & \textbf{1.68} & \textbf{0.6120} & 10.78 & \textbf{0.6834} & \textbf{17.40} & 0.6427 & \textbf{20.16} & 0.3999 \\
    \textbf{+ perp} & 17.07 & \textbf{0.5950} & 3.17 & \textbf{0.6127} & 13.85 & \textbf{0.6943} & \textbf{18.65} & \textbf{0.6562} & \textbf{17.07} & 0.4540 \\
    \textbf{+ alpha} & \textbf{12.67} & 0.5895 & \textbf{0.81} & \textbf{0.6016} & \textbf{10.55} & 0.6832 & \textbf{14.96} & \textbf{0.6440} & 27.03 & \textbf{0.4916} \\
    \textbf{+ CL} & 16.86 & \textbf{0.6150} & \textbf{2.16} & 0.6011 & \textbf{7.72} & \textbf{0.6924} & 19.45 & \textbf{0.6644} & 22.59 & \textbf{0.4840} \\
    \hline
  \end{tabular}
  \caption{The results of FSIG on FFHQ $\to$ Amedeo, Raphael, Otto; LSUN Church $\to$ Haunted; and LSUN Car $\to$ Wrecked. We report KID$\times 10^3$@5k, and Intra-cluster LPIPS@1k. Top-3 results are in bold.}
  \label{tab:quan2}
\end{table*}

\begin{figure*}[p]
  \centering
  \includegraphics[width=1.0\linewidth]{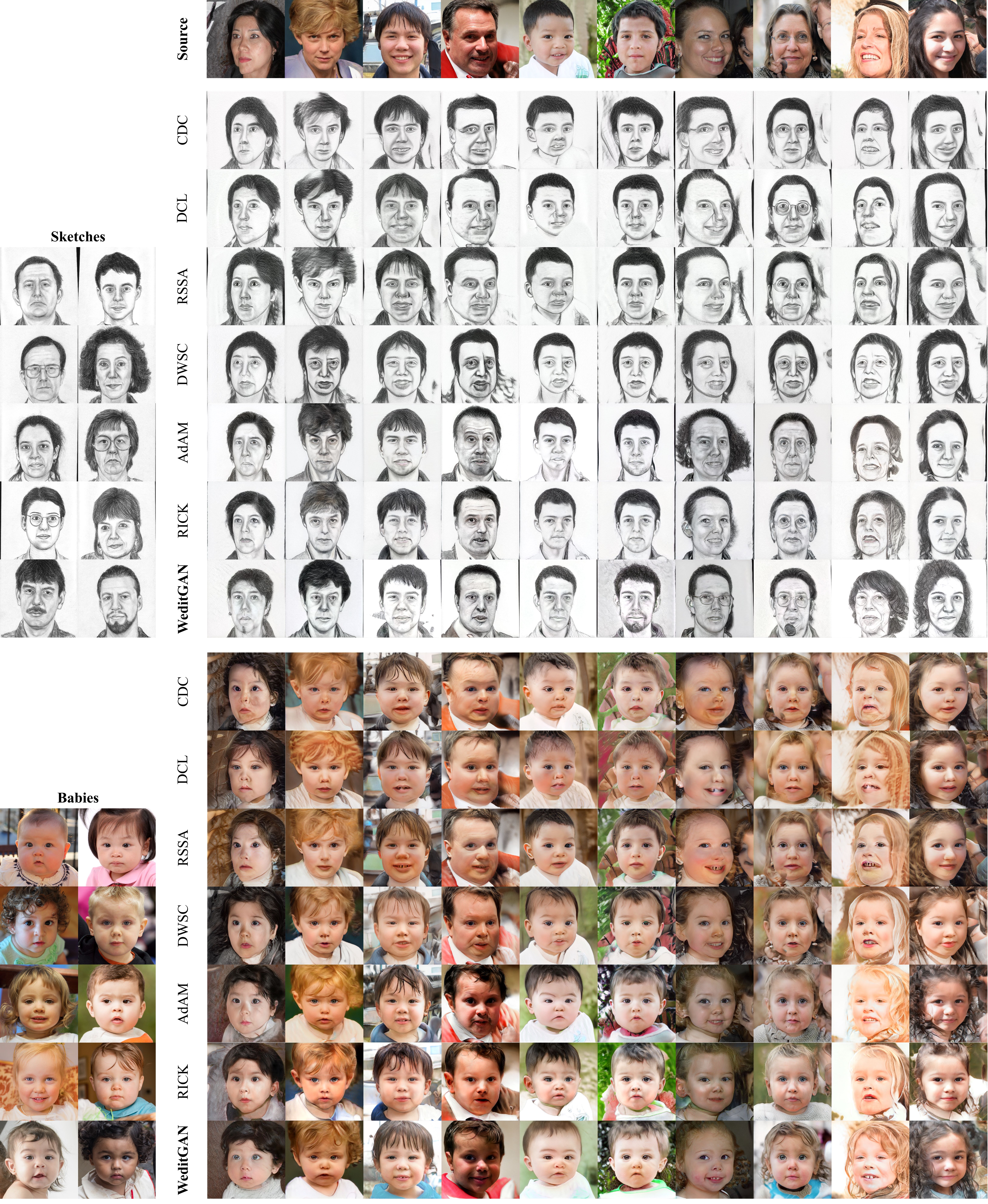}
  \caption{The 10-shot datasets (left), the generated samples of source domain FFHQ (top), target domain Sketches (middle) and Babies (bottom). Generated samples in each column are generated with the same random input $z$.}
  \label{fig:qual}
\end{figure*}

\paragraph{Qualitative Result}

We depict the generated samples of Sketches and Babies in \cref{fig:qual}. See \supp{} for the other domains, and \cref{sec:ablation} for visual comparisons between \name{} and its variants.

As the generated samples show, \name{} achieves good balance between inheriting the characteristics of the source domains and capturing those of the target domains. The generated images not only roughly share similar attributes (\eg poses, appearances, expressions) with their corresponding source images from $\wsrc$, but also present the symbolic attributes of the target domains. In Sketches, our images have similar artistic styles of strokes, shadows, and other facial details with the dataset images. In Babies, \name{} learns to produce round faces with large eyes and small noses look natural to infants. With these iconic characteristics covered, \name{} attains competitive image fidelity, while still keeping high diversity. Such observation matches the quantitative results in \cref{tab:quan1}.

Among the baselines, many images rendered by the recent finetuning-based methods AdAM and RICK show unnatural distortions and artifacts. As for the regularization-based methods CDC, DCL, RSSA and DWSC, their images manifest low diversity in local semantic regions (\eg eyes, noses, mouths) possibly due to the usage of patch discriminators. 

Sometimes the samples generated by regularization-based methods  seem more similar to the corresponding source images than the finetuning-based methods (including \name{}). The direct reason is that they impose strong regularization terms penalizing feature changes, which encourage preserving the attributes of the source images. 
However, such regularization may hinder the model from capturing target characteristics when such characteristics conflict with the source domain. 
For example, these methods try to produce small and round baby faces while keeping larger and longer adult faces, making the faces obscure and unnatural.
Since the ultimate goal of FSIG is to generate images faithful to the target domain, high similarity between source/target images should not be overemphasized.
\cref{tab:quan1} have manifested the superiority of \name{} over these regularization-based methods with quantitative evaluation objectively without the involvement of the source images.

\begin{figure*}[t]
  \centering
  \includegraphics[width=\textwidth]{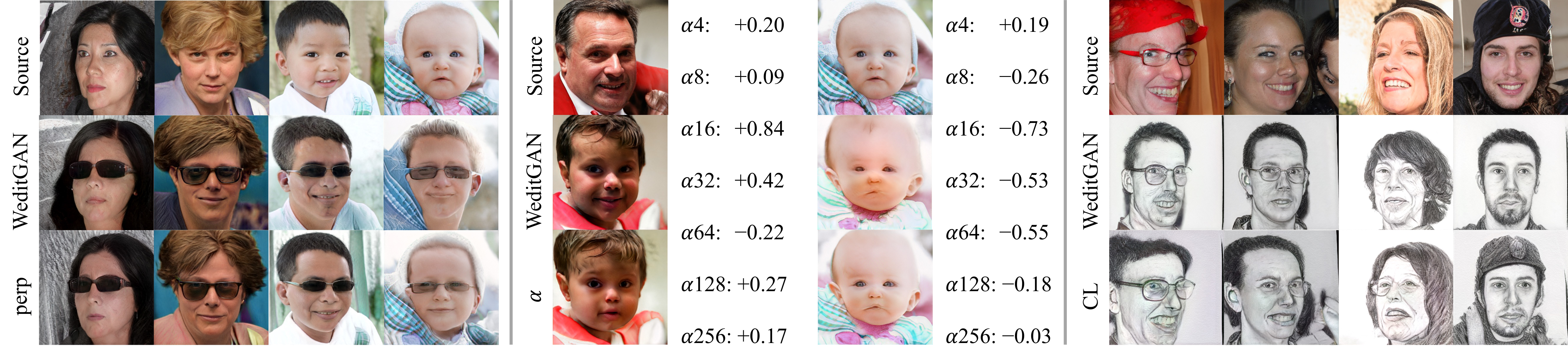}
   \caption{Visual comparisons between \name{} and its three variants. Left: \name{} perp on Sunglasses. Middle: \name{} alpha on Babies. $\alpha4, \alpha8, \dots, \alpha256$ are the intensity residuals corresponding to the synthesis blocks at resolution $4, 8, \dots, 256$. Right: \name{} CL on Sketches.}
   \label{fig:variant}
\end{figure*}



\subsection{Ablation Study}
\label{sec:ablation}

\paragraph{\name{} vs. Variants}

In \cref{fig:variant}, we provide visual comparisons between \name{} and its variants. For \name{} perp on Sunglasses, the perpendicular regularization reduces unnecessary edits on facial appearances, hairstyles or clothes. By relocating the latent space according to $P_{\mathrm{domain}}$ rather than $P_{\mathrm{data}}$, \name{} perp generally improves the diversity in \cref{tab:quan1,tab:quan2}. For \name{} alpha on Babies, the AlphaModules strengthen the editing intensities for elders and weaken those for children, respectively preventing under-editing or over-editing. Therefore, \name{} alpha increases the image fidelity for most cases in \cref{tab:quan1,tab:quan2}. For \name{} CL equipped with contrastive losses, the model preserves more source characteristics, as source/target image pairs possess higher resemblance.

\paragraph{\name{} Designs}

\begin{table}[t]
  \centering
    \small
  \begin{tabular}{lrr}
    \hline
     & \multicolumn{2}{c}{Sketches}  \\
    Method & FID$\downarrow$ & LPIPS$\uparrow$  \\
    \hline
    \textbf{\name{}} & 35.41 & 0.4339 \\
    w/ unified const $\dw$ & 88.73 & 0.4514 \\
    w/ adaptive $\dw$ & 37.70 & 0.2515 \\
    + FreezeD 13 & 46.21 & 0.3806 \\
    + FreezeD 19 & 43.74 & 0.3837 \\
    \hline
  \end{tabular}
  \caption{The results of \name{} with other relocation methods or partially freezed discriminators on Sketches.}
    \label{tab:ablation}
\end{table}

To verify that constant $\dw$ in the extended latent space $\Wp$ is the best choice, we try other relocation designs. Firstly we use a unified constant $\dw \in \W$ for all convolutional layers instead of separate ones. \cref{tab:ablation} shows that such unified $\dw$ is not flexible enough to construct the target latent space, thus rendering low quality.

Secondly, we replace constant $\dw$ with adaptive $\dw$ learned from $\wsrc$ via small networks of two FC layers. As in \cref{tab:ablation}, adaptive $\dw$ faces overfitting due to data insufficiency. 
However, when adaptive $\dw$ is used in transferring to the relatively adequate full dataset of Sketches without obvious overfitting, the near-to-one cosine similarities among adaptive $\dw$ have inspired our final design of constant $\dw$, see \supp{} for further investigations.

We also justify that finetuning the whole discriminator will not cause diversity degradation in \name{}. We combine \name{} with FreezeD \cite{freezed}, a popular strategy to finetune discriminators. FreezeD 13/19 fixes the lowest 13/19 layers of the discriminiator, and only the remaining layers are trainable.\cref{tab:ablation} shows that finetuning the full discriminator does not result in overfitting.

\section{Conclusion}
\label{sec:conclusion}

We propose \name{}, a finetuning-based FSIG method. By relocating the latent spaces with learned constant latent offsets, \name{} is able to transfer the model pretrained on large source domain to few-shot target ones. Compared to previous works, our method balances well between source/target domains, generating images both inheriting the diversity of the source domain and faithful to the target domain. The experimental results of \name{} and its variants also manifest their capability as a simple yet highly effective method. Possible limitations and future works are discussed in \supp{}.

\clearpage

\section*{Ethics Statement}
Depending on the specific applications, possible societal harms of the few-shot image generation method proposed in this work can be (1) generating fake images for misuse; and (2) copyright violation. The authors hereby solicit proper usage of this work.

\section*{Acknowledgements}
This work was supported by the Shanghai Municipal Science and Technology Major/Key Project, China (Grant No. 2021SHZDZX0102, Grant No. 20511100300) and the National Natural Science Foundation of China (Grant No. 62076162).  

\bibliography{weditgan}

\clearpage
\appendix
\section*{\huge Appendix}

As the supplementary material for \emph{WeditGAN: Few-Shot Image Generation via Latent Space Relocation}, we will clarify main updates of this version in \cref{sec:version}, list implementation details in \cref{sec:supp_detail}, show additional experimental results in \cref{sec:supp_exp}, and describe possible limitations and potential future works based on \name{} in \cref{sec:supp_future}.

\section{Update Notes}
\label{sec:version}

Compared with the previous version (v2), this version (v3) mainly updates the following contents.
\begin{itemize}
    \item Ethics Statement and Acknowledgements have been attached to the main paper.
    \item Codes have been released at \url{https://github.com/Ldhlwh/WeditGAN}.
    \item Other minor modifications.
\end{itemize} 
\section{Implementation Detail}
\label{sec:supp_detail}

\paragraph{Network Structure} \name{} is based on the official GitHub repository of StyleGAN2-ADA \cite{ada} provided by NVIDIA\footnote{\url{https://github.com/NVlabs/stylegan2-ada-pytorch}}. Since we do not modify either the two networks (mapping and synthesis) of the generator or the discriminator, our network structures are identical to those of aforementioned implementation.

For the models pretrained on FFHQ \cite{stylegan1} or LSUN Church \cite{lsun}, we choose \emph{auto} configuration of the official StyleGAN2-ADA implementation whose mapping network consists of two FC layers, using four GPUs. For the model pretrained on LSUN Car, we use \emph{paper256\_4gpu} modified from \emph{paper256} configuration using four GPUs instead of eight as the only difference. This configuration adopts an eight-layer mapping network since \emph{auto} does not render good performance on the noisy LSUN Car dataset.

Some previous works use a third-party repository of StyleGAN2\footnote{\url{https://github.com/rosinality/stylegan2-pytorch}}, \eg \citet{cdc, dcl, rick}. This implementation by default uses eight FC layers in its mapping network and doubles the convolutional channels in synthesis blocks at resolution 64--256 (by setting channel multiplier to 2). For fair comparisons, \name{} and all the baselines should perform model transfers from the same pretrained models. Hence we stick to the original NVIDIA StyleGAN2 implementation described above and slightly modify the codes of these previous works \wrt network structure.

\paragraph{Training Protocol}

The pretrained StyleGAN models have FID@50k scores of $4.07$, $3.96$ and $5.76$ respectively on FFHQ, LSUN Church and LSUN Car, after training for 23k, 48k and 83k kimgs on four GPUs. During the transfer process, \name{} and its variants are trained for 400 kimgs on a single GPU, with random seed set as 0 by default.

\paragraph{Evaluation Method}

All the metrics used in our work (FID, KID and Intra-cluster LPIPS) are operated on the features extracted by pretrained feature encoders. In practical, the choice of different encoders will significantly affect the scores of these metrics. We use the pretrained Inception-v3 provided in TorchVision\footnote{\url{https://pytorch.org}} for FID/KID, and VGG for LPIPS\footnote{\url{https://github.com/richzhang/PerceptualSimilarity}}, following previous works.

\paragraph{Simplification of Perpendicular Loss}

In the variant of WeditGAN perp, the perpendicular loss in \cref{eq:perp2} involves a nested loop to compute the inner product of $\dw$ and the difference of each $(\wsrc^i, \wsrc^j)$ pair. We may further simplify \cref{eq:perp2} into
\begin{equation}
    \Lperp = \sum_{1 \le i \le m - 1} \langle \dw, \wsrc^i - \wsrc^{i + 1} \rangle^2,
    \label{eq:perp3}
\end{equation}
since we have 
\begin{equation}
    \dw \perp \wsrc^i - \wsrc^j \Leftrightarrow \dw \perp \wsrc^j - \wsrc^i
\end{equation}
and
\begin{equation}
    \begin{aligned}
    & \left\{
        \begin{aligned}
            \dw \perp \wsrc^i - \wsrc^j \\
            \dw \perp \wsrc^j - \wsrc^k 
        \end{aligned}\right. \\
    & \Rightarrow \dw \perp \wsrc^i - \wsrc^k.
    \end{aligned}
\end{equation}
In practical, we use the simplified perpendicular loss in \cref{eq:perp3} instead of \cref{eq:perp2} to accelerate computation.

\paragraph{Code}

Codes of WeditGAN have been released at \url{https://github.com/Ldhlwh/WeditGAN}.

\section{Additional Result}
\label{sec:supp_exp}

First we detail the preliminary experiments with adaptive $\dw$ in \cref{sec:adaptdw}. Second we report the training time and the numbers of trainable parameters of \name{} and the baselines in \cref{sec:timeandparam}. Next we depict generated images with interpolation to show the interpretable meanings of the learned $\dw$ in \cref{sec:itp}. Then we illustrate the qualitative results of the other six target domains in \cref{sec:qual}, which are not included in our main paper. Besides, we also perform 5-shot/1-shot image generation experiments in \cref{sec:fewer} and significance tests in \cref{sec:sig}.

\subsection{Experiments with Adaptive $\dw$}
\label{sec:adaptdw}

\begin{table}[t]
  \centering
    \small
    \setlength{\tabcolsep}{1.6mm}
  \begin{tabular}{r|cc|cc|cc}
    \hline
    & \multicolumn{2}{c|}{Conv0} & \multicolumn{2}{c|}{Conv1} & \multicolumn{2}{c}{ToRGB} \\
    Res. & $\cos$ & L1 & $\cos$ & L1 & $\cos$ & L1 \\
    \hline
    4 & 0.9989 & 25.4601 & - & - & 0.9408 & 0.0463 \\
    8 & 0.9924 & 28.3873 & 0.9924 & 29.1519 & 0.9998 & 0.0003 \\
    16 & 0.9943 & 38.0310 & 0.9720 & 47.9177 & 0.8846 & 0.1769 \\
    32 & 0.9521 & 40.3618 & 0.9954 & 30.2054 & 0.9082 & 0.1527 \\
    64 & 0.9875 & 30.9129 & 0.9888 & 39.9418 & 0.9990 & 0.0004 \\
    128 & 0.9898 & 27.9454 & 0.9867 & 52.8873 & 0.9331 & 0.2027 \\
    256 & 0.9977 & 40.2286 & 0.9995 & 13.0162 & 0.9975 & 0.4360 \\
    \hline
  \end{tabular}
  \caption{The average cosine similarity and L1 distance of 1,000 adaptive $\dw$ corresponding to each layer in the synthesis network, from 1,000 randomly sampled input $z$. The results are from the model transfer experiment from FFHQ to the full dataset of Sketches.}
    \label{tab:adaptdw}
\end{table}

In \name{} with adaptive $\dw$, we learn one small network consists of two FC layers for each layer in the synthesis network, rendering the latent offset which modulates the convolutional filters of that layer. With 1,000 random $z$ as the input, we calculate the average cosine similarity and the average L1 distance of the adaptive $\dw$ corresponding to each layer in the experiment transferring from FFHQ \cite{stylegan1} to the full dataset (around 300 images) of Sketches \cite{sketches}, shown in \cref{tab:adaptdw}. For the synthesis layers (named \emph{Conv} in \cref{tab:adaptdw}) where the latent codes $w$ are normalized before modulating the convolutional filters (thus L1 distance is not important), the average cosine similarity among these 1,000 $\dw$ is almost one. While for the ToRGB layers where the latent codes are not normalized, similarity of these $\dw$ is manifested from both the cosine similarity and the L1 distance. The findings above prove that learning constant $\dw$ is a feasible choice to relocate the target latent space, especially when the data is insufficient.

\subsection{Training Time and Parameters}
\label{sec:timeandparam}

The training time to the best results and the numbers of trainable parameters during model transfer from FFHQ to Sketches with resolution $256^2$ are listed in \cref{tab:timeandparam}. For an official StyleGAN2 model (\emph{auto} configuration) containing 23.2M and 24.0M parameters in its generator and discriminator respectively, the regularization-based methods CDC \cite{cdc}, DCL \cite{dcl}, RSSA \cite{rssa} and DWSC \cite{dwsc} finetune all the parameters in both networks, with actually around 18k more for the patch discriminator. The recent finetuning-based method AdAM \cite{adam} and RICK \cite{rick} reduces the trainable parameters by applying kernel modulation, or by fixing or pruning a certain portion of the parameters. Finally, our \name{} achieves state-of-the-art performance by training just $0.04\%$ of the generator parameters. Besides, in scenarios that we need to transfer the source model to multiple target domains (say, $n$ domains). \name{} only needs to store $23.2\mathrm{M} + 10240 \cdot n$ parameters, while this number for other methods is generally $23.2\mathrm{M} \times n$. As for the training time, that of \name{} is also the shortest among the competitive baselines.

\begin{table}[t]
  \centering
  \small
  \begin{tabular}{lrrr}
    \hline
    Method & Time & \#param of $G$ & \#param of $D$ \\
    \hline
    CDC & 2h 7min & 23.2M & 24.0M \\
    DCL & 4h 51min & 23.2M & 24.0M \\
    RSSA & 3h 34min & 23.2M & 24.0M \\
    DWSC & 2h 3min & 23.2M & 24.0M \\
    AdAM & 1h 39min & 18.8M & 24.0M \\
    RICK & 1h 5min & - & - \\
    \hline 
    \textbf{\name{}} & 46min & 10,240 & 24.0M\\
    \hline
  \end{tabular}
  \caption{The training time and the number of trainable parameters of the few-shot image generation experiment on the target domain Sketches. The training time is recorded on a single NVIDIA TITAN RTX. Since in RICK the training policies of the parameters are altered several times among three modes fixed, trainable or pruned on-the-fly during the transfer processes, the number of trainable parameters may vary so we omit it in this table. However most of the parameters are trainable at least for a certain period.}
    \label{tab:timeandparam}
\end{table}

\subsection{Inspection of the Learned $\dw$}
\label{sec:itp}

\begin{figure}[t]
  \centering
  \includegraphics[width=1.0\linewidth]{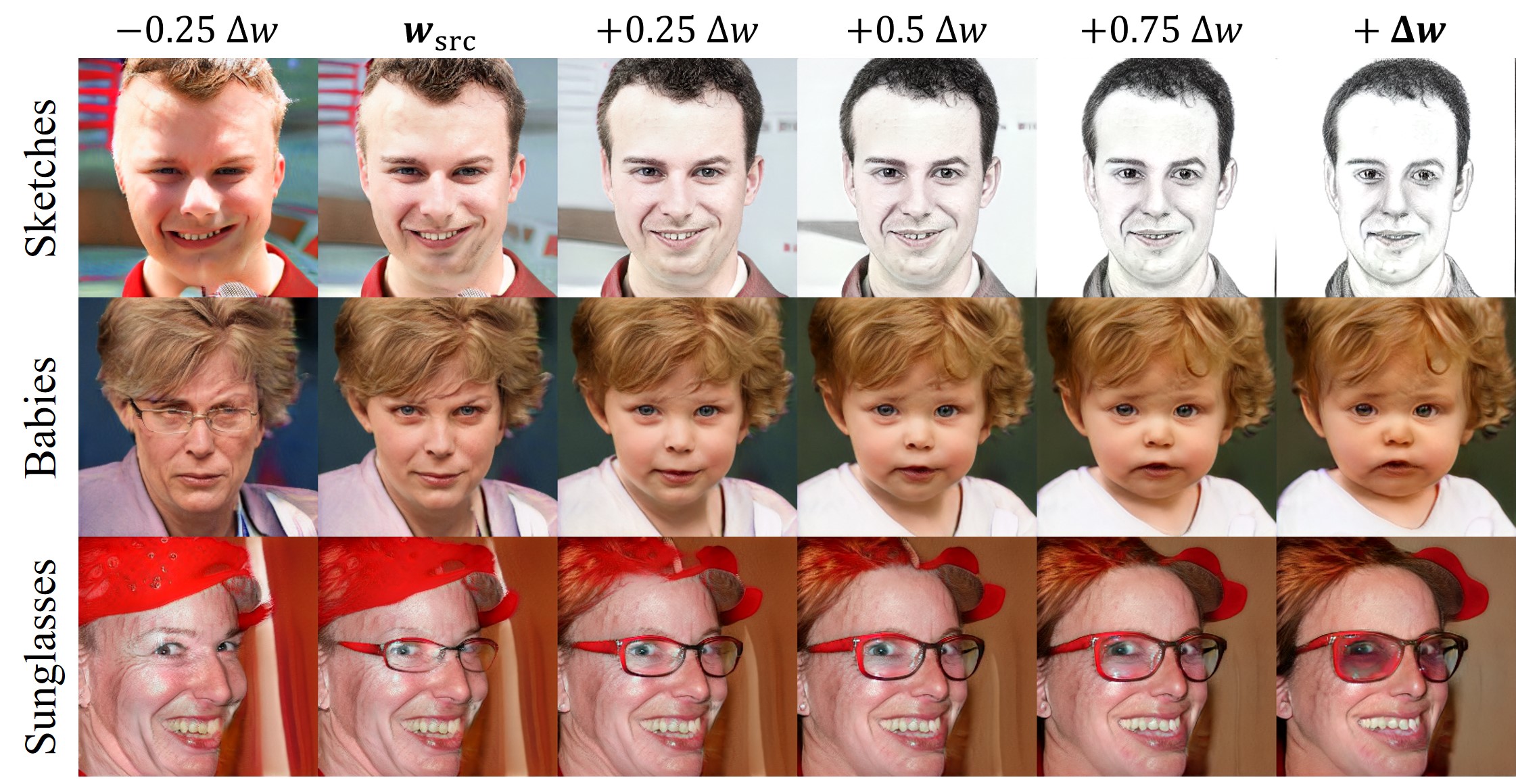}
  \caption{Samples generated by \name{} using $\dw$ learned on Sketches, Babies or Sunglasses with different intensities ranging from $-0.25$ to $1.0$.}
  \label{fig:itp}
\end{figure}

We review whether the learned $\dw$ holds interpretable meanings bridging the gap between the source/target domains. Therefore, we generate images using $w = \wsrc + \lambda \cdot \dw$, where $\lambda$ ranges from $-0.25$ to $1.0$. As the generated samples with $\dw$ learned on three target domains depicted in \cref{fig:itp}, ascending positive $\lambda$ can smoothly transfer source images to target ones as expected. Also, negative $\lambda$ can reverse the targeting attributes. In Sketches, the person in the image with reversed $\dw$ is stouter and has saturated color since those in Sketches are generally thinner and in grayscale. In Babies, the person becomes aged with longer face. In Sunglasses, even the glasses in the source image are removed. These examples validate that $\dw$ learned by \name{} are meaningful, and worth deeper inspection in future works.

\begin{figure*}[p]
  \centering
  \includegraphics[width=1.0\linewidth]{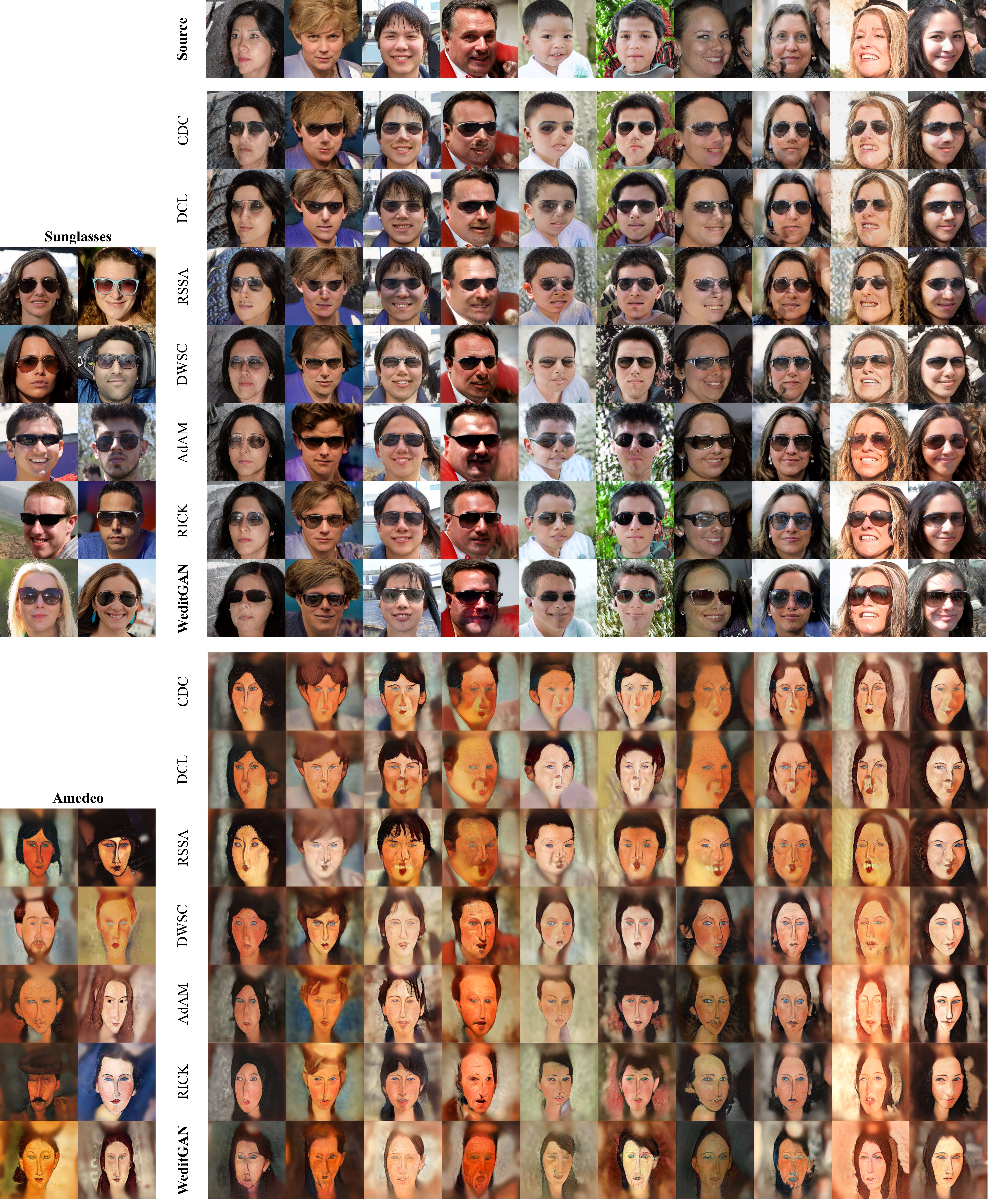}
  \caption{The 10-shot datasets (left), the generated samples of source domain FFHQ (top), target domain Sunglasses (middle) and Amedeo (bottom). Generated samples in each column are generated with the same random input $z$.}
  \label{fig:qual1}
\end{figure*}
\begin{figure*}[p]
  \centering
  \includegraphics[width=1.0\linewidth]{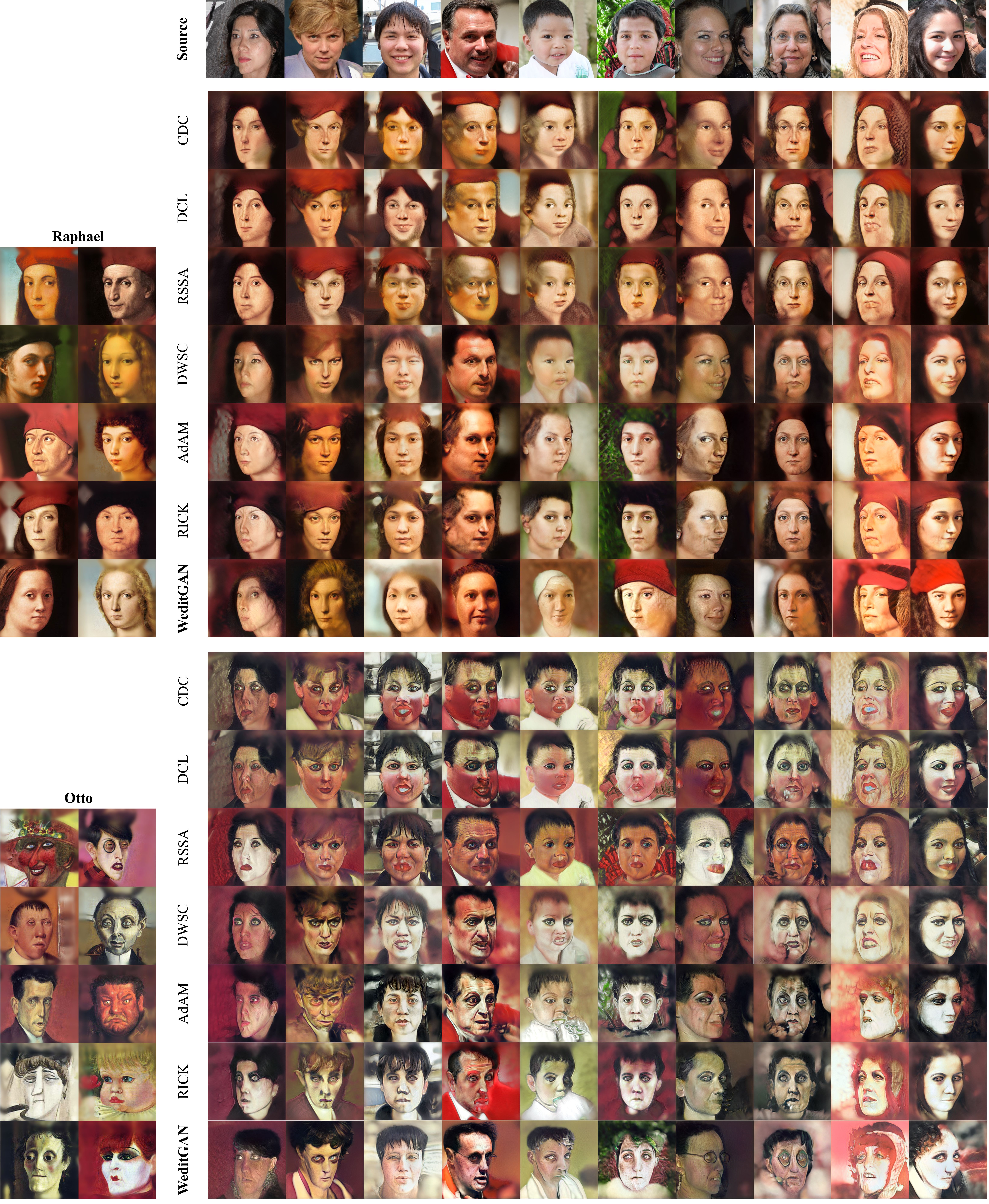}
  \caption{The 10-shot datasets (left), the generated samples of source domain FFHQ (top), target domain Raphael (middle) and Otto (bottom). Generated samples in each column are generated with the same random input $z$.}
  \label{fig:qual2}
\end{figure*}
\begin{figure*}[p]
  \centering
  \includegraphics[width=0.92\linewidth]{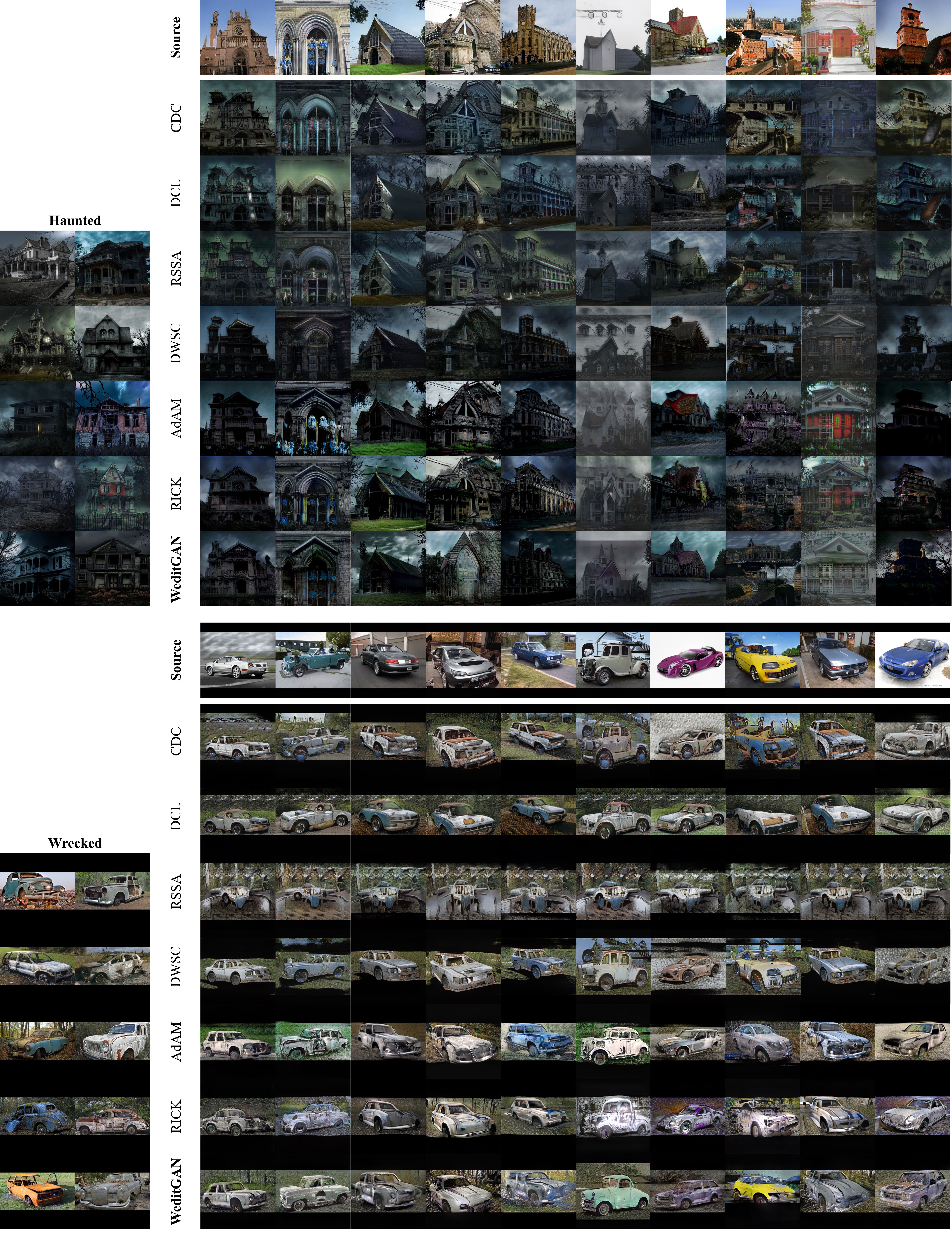}
  \caption{The 10-shot datasets (left), the generated samples of source domain LSUN church, target domain Haunted, source domain LSUN car and target domain Wrecked (from top to bottom). Generated samples in each column are generated with the same random input $z$.}
  \label{fig:qual3}
\end{figure*}

\subsection{Qualitative Result}
\label{sec:qual}

The dataset images and the generated samples of FFHQ $\to$ Sunglasses \cite{stylegan1}, Amedeo, Raphael, Otto \cite{faceofart}, and those of LSUN church $\to$ Haunted and LSUN car $\to$ Wrecked \cite{cdc} are shown in \cref{fig:qual1,fig:qual2,fig:qual3}. In Sunglasses, our glasses are realistic with few artifacts. In Amedeo, \name{} captures the narrow faces and tall noses. In Raphael, the faces generated by \name{} are the most natural. In Otto, our method produces a variety of facial appearances (\eg eyes, mouths) compared with previous works. In Haunted, our houses possess pitched roofs as a symbolic characteristic commonly seen in the dataset images, and so are the looks of classic cars in Wrecked.  These images demonstrate the uniformly satisfying performances of our \name{}, whose generated samples capture the key attributes of the target domains.

\begin{table}[t]
  \centering
  \small
  \begin{tabular}{lrr|rr}
    \hline
     & \multicolumn{2}{c|}{Sketches 5-shot} & \multicolumn{2}{c}{Sketches 1-shot}  \\
    Method & FID$\downarrow$ & LPIPS$\uparrow$ & FID$\downarrow$ & LPIPS$\uparrow$  \\
    \hline
    CDC & 52.67 & 0.3831 & 68.24 & 0.3313 \\
    DCL & 62.23 & 0.4139 & 73.43 & 0.4364 \\
    RSSA & 62.93 & 0.4439 & 68.69 & 0.3798 \\
    DWSC & 66.77 & 0.3270 & 83.23 & 0.3733 \\
    AdAM & 42.66 & 0.4098 & 80.83 & 0.3411 \\
    RICK & 41.60 & 0.3888 & 71.91 & 0.4347 \\
    \hline 
    \textbf{\name{} perp} & 37.85 & 0.4297 & 66.02 & 0.4391 \\
    \hline
  \end{tabular}
  \caption{The results of 5-shot/1-shot image generation experiments on FFHQ $\to$ Sketches. We report FID@5k with the full dataset, and Intra-cluster LPIPS@1k. Note that in 1-shot setting Intra-cluster LPIPS regresses to standard LPIPS.}
    \label{tab:fewer}
\end{table}

\begin{figure*}[p]
  \centering
  \includegraphics[width=0.92\linewidth]{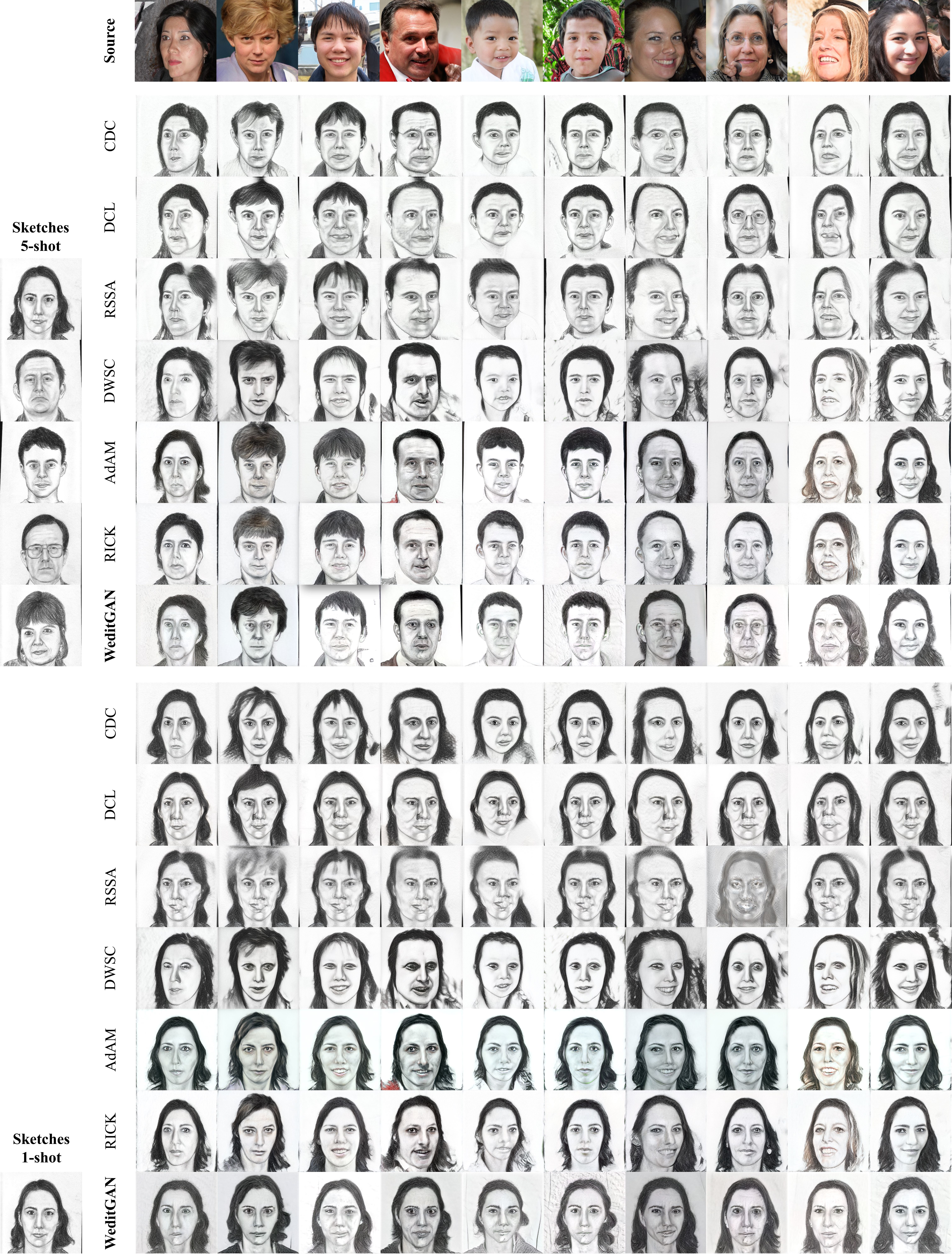}
  \caption{The datasets (left) and the generated samples of source domain FFHQ (top), target domain Sketches in 5-shot (middle) and 1-shot (bottom). Generated samples in each column are generated with the same random input $z$.}
  \label{fig:fewer}
\end{figure*}

\subsection{Fewer-shot Image Generation}
\label{sec:fewer}

To verify the generative ability of our \name{} in extreme few-shot image generation cases, we redo the experiments on target domain Sketches with only five or one images available for training. These experiments share the same settings with the 10-shot ones in the main paper except for the numbers of training images. We select the variant of \name{} with perpendicular regularization since the gap between $P_{\mathrm{data}}$ and $P_{\mathrm{domain}}$ is even greater in these extremely data-insufficient cases.

The quantitative and the qualitative results are shown in \cref{tab:fewer} and \cref{fig:fewer}. Our method achieves competitive scores among the recent works. Besides, compared with the 10-shot experiments in the main paper, the performance of \name{} does not deteriorate much in the 5-shot case probably because five images are enough to roughly locate the target latent space given the shape of its manifold we borrow from the source latent space. In the 1-shot case, our \name{} still manages to maintain the overall quality and the variety of facial details to certain extent by producing different eyes, noses, face shapes, \etc.

\subsection{Significance Test}
\label{sec:sig}

To ensure that our \name{} is significantly superior to the previous works in the statistical point of view, we do significance tests on Sketches between the latest baseline RICK \cite{rick} and \name{} perp. We repeat the 10-shot image generation experiments ten times using different random seeds. According to the quantitative results in \cref{tab:sig}, we conduct Welch's $t$-test on both metrics. The $p$-value on FID and Intra-cluster LPIPS are respectively $1.91\times 10^{-12}$ and $6.66 \times 10^{-8}$, which proves the statistically significant superiority of our method.

\begin{table}[t]
  \centering
  \small
  \begin{tabular}{lcc}
    \hline
    Method & FID$\downarrow$ & LPIPS$\uparrow$ \\
    \hline
    RICK & 39.99 $\pm$ 0.37 & 0.4301 $\pm$ 0.0049 \\
    \textbf{\name{} perp} & 37.05 $\pm$ 0.25 & 0.4510 $\pm$ 0.0052 \\
    \hline
  \end{tabular}
    \caption{The results of significance tests on target domain Sketches, where we report Mean$\pm$Std of ten times.}
    \label{tab:sig}
\end{table}
\begin{figure*}[t]
  \centering
  \includegraphics[width=.9\linewidth]{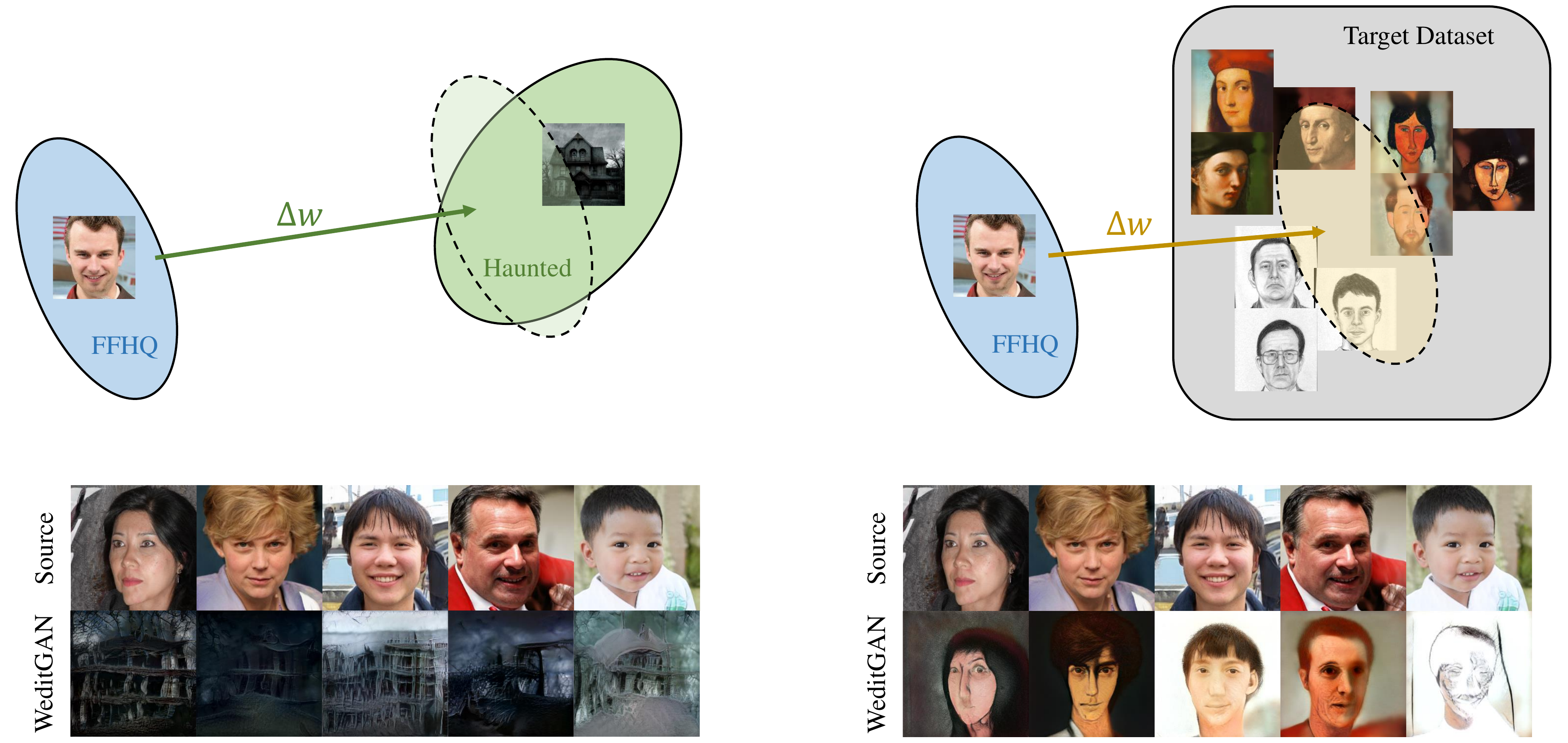}
   \caption{Limitations of \name{}. Left (FFHQ $\to$ Haunted): The source/target domains should be depicting the same category of objects, otherwise the source/target latent spaces may not have the same shapes of manifolds within the same model, and the synthesis network is not well-trained on generating other categories of objects either. Right (FFHQ $\to$ mixed dataset of Sketches, Amedeo and Raphael): The target datasets should have unified styles (\ie from single domains), otherwise \name{} may fail to cover multiple domains. The generated images may have fused characteristics of these domains.}
   \label{fig:limit}
\end{figure*}

\section{Limitation and Future Work}
\label{sec:supp_future}

Objectively speaking, though our \name{} renders satisfying performance in the few-shot image generation experiments, a few limitations still indicate possible improvement worthy of future research efforts.

First, as shown in \cref{fig:limit} (left), \name{} assumes similar shapes of manifolds of the source/target latent spaces, hence pairs of source/target domains have to be structurally or semantically related. The source/target domains should focus on the same category of objects (\eg FFHQ and Sketches both depict human faces). If the source/target datasets focus on different categories of objects (\eg FFHQ $\to$ Haunted) thus possess significantly different distributions, it can be foreseen that \name{} cannot construct the target latent space only with a constant $\dw$. Furthermore, the synthesis network of the generator pretrained on one category of objects may not be able to generate images of other categories without finetuning.

Second, as shown in \cref{fig:limit} (right), since \name{} only learns one constant $\dw$ during the transfer process, the images in a target dataset should be of a unified style (\eg portraits painted by the same artist). If the target dataset consists of images from multiple domains with different styles (\eg a collection of portraits by several artists), our method may fail to cover these domains with only a single $\dw$.

Despite the aforementioned limitations, it is usually deemed reasonable to do model transfer only between related domains, where neither of the source/target datasets contains images from multiple domains. Therefore, our method is practically capable in most real-world cases. 
In conclusion, we hope that our research can serve as a pivotal work to inspire future methods based on latent manipulation in the area of few-shot image generation.

\end{document}